\documentclass{article}

\usepackage{PRIMEarxiv}
\usepackage{authblk}
\usepackage[utf8]{inputenc} 
\usepackage[T1]{fontenc}    
\usepackage{hyperref}       
\usepackage{url} 
\usepackage{bm}
\usepackage{booktabs}       
\usepackage{multirow}
\usepackage{graphicx}
\usepackage{amsfonts}       
\usepackage{nicefrac}       
\usepackage{microtype}      
\usepackage{lipsum}
\usepackage{fancyhdr}       
\usepackage{graphicx}       
\graphicspath{{media/}}     
\usepackage{amsmath}
\usepackage{dsfont}

\usepackage{comment}
\title{Distribution-Free Uncertainty-Aware Virtual Sensing via Conformalized Neural Operators
}

\author[1,2*]{Kazuma Kobayashi}
\author[3*]{Shailesh Garg}
\author[2]{Farid Ahmed}
\author[1,3,4]{Souvik Chakraborty}
\author[1,2]{Syed Bahauddin Alam}

\affil[1]{The Grainger College of Engineering, Nuclear, Plasma \& Radiological Engineering, University of Illinois Urbana-Champaign, Urbana, IL, USA
}

\affil[2]{National Center for Supercomputing Applications, Urbana, IL, USA
}

\affil[3]{Department of Applied Mechanics, Indian Institute of Technology Delhi, New Delhi, India
}

\affil[4]{Yardi School of Artificial Intelligence, Indian Institute of Technology Delhi}

\begin{document}
\maketitle

\begin{abstract}

Robust uncertainty quantification (UQ) remains a critical barrier to the safe deployment of deep learning in real-time virtual sensing, particularly in high-stakes domains where sparse, noisy, or non-collocated sensor data are the norm. We introduce the Conformalized Monte Carlo Operator (CMCO), a framework that transforms neural operator-based virtual sensing with calibrated, distribution-free prediction intervals. By unifying Monte Carlo dropout with split conformal prediction in a single DeepONet architecture, CMCO achieves spatially resolved uncertainty estimates without retraining, ensembling, or custom loss design. Our method addresses a longstanding challenge: how to endow operator learning with efficient and reliable UQ across heterogeneous domains. Through rigorous evaluation on three distinct applications—turbulent flow, elastoplastic deformation, and global cosmic radiation dose estimation—CMCO consistently attains near-nominal empirical coverage, even in settings with strong spatial gradients and proxy-based sensing. This breakthrough offers a general-purpose, plug-and-play UQ solution for neural operators, unlocking real-time, trustworthy inference in digital twins, sensor fusion, and safety-critical monitoring. By bridging theory and deployment with minimal computational overhead, CMCO establishes a new foundation for scalable, generalizable, and uncertainty-aware scientific machine learning.

\end{abstract}


\section{Introduction}
Artificial intelligence (AI) has been increasingly used in engineering domains to support the modeling, predicting, and controlling of complex physical systems. In energy applications, AI has been applied to tasks such as power plant operation monitoring, component diagnostics, and forecasting the remaining useful life (RUL) \cite{kobayashi2024explainable}, where accurate field estimation is critical for ensuring system safety, reducing operational costs, and enhancing overall performance.

One particularly important use of AI in these applications is virtual sensing \cite{kobayashi2025proxies,kobayashi2024virtual,hossain2025virtual}, in which unmeasured physical quantities are inferred from limited, indirect, or noisy sensor inputs. This approach is valuable in scenarios where direct measurements are infeasible due to spatial inaccessibility, sensor failure, cost constraints, or safety considerations \cite{hossain2025virtual,kobayashi2024virtual,kobayashi2024deep}. Virtual sensing enables various capabilities, including estimating boundary conditions, predicting internal system states, reconstructing missing or corrupted data, and monitoring safety-critical variables that cannot be directly measured. In many cases, estimating these quantities at a few key locations may be sufficient. However, more demanding scenarios may require full-field reconstruction, such as safety-critical systems or high-fidelity digital twins, to recover high-dimensional, spatially distributed fields from sparse observations. By enabling such capabilities, virtual sensing significantly enhances observability, situational awareness, and real-time decision-making in complex engineering systems.

To support such inference tasks, neural operator learning has emerged as a powerful framework for mapping between infinite-dimensional function spaces. It offers significant advantages over traditional neural networks in solving parametric partial differential equations (PDEs) and modeling complex system dynamics. Several classes of neural operators have been developed, including the Fourier Neural Operator (FNO) \cite{li2020fourier,bonev2023spherical}, Graph Neural Operator (GNO) \cite{li2020neural,li2020multipole}, and Wavelet Neural Operator (WNO) \cite{tripura2023wavelet_mech,tripura2023wavelet}. These models have demonstrated strong performance in learning solution operators over structured and unstructured domains. The FNO performs spectral convolution in the Fourier domain, allowing efficient learning of translation-invariant dynamics. The GNO extends operator learning to graph-structured inputs, while the WNO incorporates multiresolution representations via wavelet transforms to capture localized features.
Despite these advancements, each model poses challenges when applied to real-world sensing and monitoring scenarios. The FNO requires complete, structured grid data and struggles to generalize in the presence of sparse, nonuniform, or missing inputs. The GNO, although capable of handling irregular topologies, depends on a predefined and fixed graph structure, which may not hold in dynamic or evolving systems. The WNO, while offering spatial adaptivity, remains constrained by spectral design assumptions and relies on consistent data structures. These limitations become particularly pronounced in practical applications involving noisy signals, incomplete sensor coverage, or non-collocated input-output domains, which are frequently encountered in engineering systems.

The Deep Operator Network (DeepONet)\cite{lu2021learning} was introduced as a general-purpose framework for learning nonlinear operators, grounded in the theory of universal operator approximation \cite{chen1995universal}. Although it was not specifically designed to address the limitations of spectral or graph-based neural operators, its design employs two sub-networks: a branch network to encode input functions and a trunk network to encode query coordinates, which naturally confers several advantages in real-world applications. A key strength of DeepONet is its flexibility in processing heterogeneous and multimodal sensor inputs \cite{jin2022mionet}. Time-series signals \cite{he2024sequential,he2024predictions,park2025bridging,park2025bridging}, spatial profiles, and image-based observations \cite{mei2024fully,huang2024porous} can be encoded as functional inputs through the branch network, enabling seamless integration of diverse sensing modalities. In addition, DeepONet supports inference at arbitrary spatial or temporal locations, handles irregular and sparse input sampling, and does not require fixed meshes or predefined graph structures. These properties make it especially well-suited for deployment in dynamic environments with disjoint input-output domains. They allow it to operate at varying spatial resolutions within a given geometry, thereby reducing the need for mesh-specific retraining and simplifying model implementation.

While neural operators have demonstrated strong capabilities in supporting virtual sensing tasks, they must also provide reliable and calibrated uncertainty quantification (UQ) when deployed in safety-critical or high-stakes applications \cite{foutch2025ai,kumar2025degradation,kobayashi2024ai,kobayashi2023practical,kobayashi2023uncertainty,kumar2021quantitative,kumar2022multi}. In such settings, it is insufficient to produce accurate predictions alone; models must also convey their confidence in those predictions, particularly when faced with sparse inputs, distributional shifts, or unobserved operating conditions.

Conventional UQ methods in deep learning, including Bayesian neural networks \cite{goan2020bayesian,jospin2022hands} and deep ensembles \cite{lakshminarayanan2017simple,ovadia2019can}, have been studied but are often ill-suited for operator-based models. Bayesian approaches typically require posterior approximations over model weights or function spaces, introducing high computational cost and sensitivity to prior assumptions. Deep ensembles require training multiple independent models, which increases memory usage and computational burden. Quantile regression is more efficient but lacks formal coverage guarantees without additional calibration and may produce inconsistent or overlapping quantile estimates.

These challenges are further amplified in neural operator models, which approximate mappings between functions and typically produce high-dimensional, spatially distributed outputs. Such models are often required to make predictions at arbitrary spatial or temporal locations, including regions not encountered during training. In many real-world sensing applications, input and output data are non-collocated. For example, internal fields may need to be inferred from limited boundary measurements. Estimating spatially resolved uncertainty under these conditions remains particularly challenging.

Recent advances have begun to address the need for distribution-free UQ in neural operator models. One such effort is our prior development of the Conformalized Randomized Prior Operator (CRP-Operator) \cite{garg2025distribution}, which combined randomized prior networks with conformal prediction to produce marginally valid uncertainty estimates without assuming Gaussianity or requiring retraining. This method used the WNO as the backbone and demonstrated effective coverage calibration on structured datasets. However, it required training multiple ensemble members with randomized initializations, which increased training costs. This work proposes a more efficient and deployment-oriented framework: the Conformalized Monte Carlo dropout Operator (CMCO). This method approximates predictive distributions using a single neural operator (i.e., DeepONet) architecture with Monte Carlo dropout at inference time. Split conformal prediction is then applied to calibrate prediction intervals at each spatial query point. The framework is model-agnostic, introduces minimal computational overhead, and avoids specialized training procedures. 

This study builds on that foundation and evaluates the proposed framework across a series of physically grounded virtual sensing tasks, progressing from canonical benchmarks to more realistic applications. These include turbulent flow reconstruction in a lid-driven cavity, stress field prediction in elastoplastic solids, and cosmic radiation dose estimation from sparse sensor inputs. Each task differs in spatial complexity and physical characteristics but presents the common challenge of reconstructing high-dimensional fields from limited or indirect measurements.

\section{Conformalized Monte Carlo Dropout Operator}
\subsection{DeepONet}
Deep Operator Network (DeepONet), introduced in \cite{lu2021learning}, is designed to learn an operator $ \mathcal{G} $ that maps an input function space $ \mathcal{I} $ to an output function space $ \mathcal{O} $. For an input function $\bm{u} \in \mathcal{I}$, the output $\mathcal{G}(\bm{u})$ is a function evaluated at location $r \in \Omega \subset \mathbb{R}^d$, i.e., $ \mathcal{G}(\bm{u})(r) \in \mathbb{R}$. 

DeepONet learns this mapping using a two-network architecture as shown in Figure \ref{fig:framework} (a). The \textit{branch net} receives the input function $\bm{u}$ evaluated at a fixed set of sensor locations $\{x_j\}_{j=1}^{n} \subset \Omega$, producing an output vector $b(\bm{u}) \in \mathbb{R}^q$. The \textit{trunk net} takes a query location $ r \in \Omega$ as input and outputs a vector $ t(r) \in \mathbb{R}^q$. The final prediction at query point $r$ is computed as the inner product:
\begin{equation}
    \mathcal{G}_\theta(\bm{u})(r) = b(\bm{u}) \cdot t(r) \simeq \mathcal{G}(\bm{u})(r),
\end{equation}
where $\theta$ denotes the trainable parameters of both networks.

This branch-trunk architecture also paves the way for specialized variants of DeepONet to specific data structures. In particular, when the input function $\bm{u}$ varies over time, such as in dynamical systems governed by temporal evolution, sensor data streams, or control inputs recorded over sequential intervals, capturing the temporal correlations inherent in the data becomes important. Sequential DeepONet (S-DeepONet) \cite{he2024sequential,he2024predictions,kobayashi2025proxies} extends the original architecture by equipping the branch network with a recurrent structure to address this. Using models such as gated recurrent units (GRUs) or long short-term memory (LSTM) networks, S-DeepONet can encode sequential dependencies in the input function before projecting it to the operator space.

This extension makes S-DeepONet particularly suitable for applications involving real-time monitoring, control, and anomaly detection in dynamic environments, where temporal coherence and history-dependent patterns play a central role in the system's behavior. Throughout this study, we adopt S-DeepONet as the core operator learning model.

\subsection{Probabilistic with Monte Carlo Dropout}
In its vanilla form, the DeepONet architecture produces deterministic outputs that convey no information regarding the uncertainty associated with the model's output. The uncertainty can originate from a lack of sufficient training data, the model's inherent limitation to learn the dataset, and an improper training algorithm. To remedy this, we utilize Monte Carlo Dropout (MC-dropout), which is an approximate Bayesian technique for quantifying uncertainty associated with a deep learning model's predictions. MC-dropout combines the MC sampling and dropout regularization, wherein the network architecture is modified to include dropout layers that mask the outputs of certain neurons at the time of training to get a regularization effect. Now, when deploying vanilla dropout layers, during inference, all neurons are kept active, i.e., no dropout masks are used as shown in Figure \ref{fig:framework} (a). However, in MC-Dropout depicted in Figure \ref{fig:framework} (b) , during inference, multiple forward passes are made for the same input while keeping the dropout mask active. This results in varied outputs for the same input, depending on which neurons are kept active in any particular forward pass. 

MC-dropout, when used within the DeepONet architecture, will produce output $\hat y_{p(i)}(u) = \mathcal G_\theta(\bm u)(y)$, for $i$\textsuperscript{th} forward pass corresponding to input function $\bm u$ and query point $y$. The final prediction of the MC-dropout DeepONet can be computed as

\begin{equation}
\mu(\bm u) = \frac{1}{n_c} \sum_{k=1}^{n_c} \hat{y}_{p(k)}(\bm u),
\label{eq:mean}
\end{equation}

\begin{equation}
\sigma(\bm u) = \sqrt{ \frac{1}{n_c} \sum_{k=1}^{n_c} \left( \mu(\bm u) - \hat{y}_{p(k)}(\bm u) \right)^2 },
\label{eq:var}
\end{equation}

where $\mu(\bm u)$ is the mean prediction and $\sigma(\bm u)$ is the standard deviation associated with the predictions. $n_c$ here is the number of forward passes at the inference stage.
As stated previously, MC-dropout is often interpreted as approximate variational Bayesian inference \cite{rasmussen2003gaussian,osband2018randomized}; it is not exact, and the accuracy of its uncertainty estimates is limited by the number of forward passes $n_c$. Therefore, the estimates obtained should be calibrated further to obtain a more accurate representation of the uncertainty associated with network predictions. For this purpose, we use split conformal prediction and introduce the CMCO framework that takes the initial estimates of uncertainty from the MC-dropout operator and calibrates them to get a more accurate uncertainty estimate.

\subsection{Conformalized Monte Carlo Dropout Operator}
Let $u \in \mathcal{U}$ denote an input function (e.g., initial or boundary condition), and let $y \in \mathbb{R}^{n_e}$ be the corresponding ground-truth solution field discretized over $n_e$ spatial locations. A total of $n$ input-output samples $\{ (u_i, y_i) \}_{i=1}^n$ are used for calibration or testing. For each input $u$, the pre-trained model is converted into an ensemble using Monte Carlo dropout during inference. Specifically, for a given test input $u_t$, a set of $n_c$ predictive samples $\{ \hat{y}_{p(k)}(u_t) \}_{k=1}^{n_c} \in \mathbb{R}^{n_e}$ is obtained through independent stochastic forward passes, where $p(k)$ indexes the $k$-th dropout realization.

The ensemble mean and standard deviation are computed elementwise using Eq.~\eqref{eq:mean} and Eq.~\eqref{eq:var}, respectively.

To construct distribution-free prediction intervals, the conformal calibration procedure proposed by Garg and Chakraborty~\cite{garg2025distribution} is adopted. Given a calibration dataset $\{ (u_i, y_i) \}_{i=1}^n$, where each $y_i \in \mathbb{R}^{n_e}$ is the ground-truth solution corresponding to $u_i$, normalized nonconformity scores are computed for each output dimension as
\begin{equation}
e_{i,j} = \frac{ \left| y_{i,j} - \mu_j(u_i) \right| }{ \sigma_j(u_i) }, \quad \text{for } i = 1, \dots, n,\; j = 1, \dots, n_e.
\end{equation}

For each output index $j$, the conformal quantile $q_j$ is computed using the $(1 - \alpha)$ quantile rule:
\begin{equation}
q_j = \text{Quantile} \left( \{ e_{1,j}, \dots, e_{n,j} \}, \frac{ \left\lceil (1 - \alpha)(n + 1) \right\rceil }{n} \right),
\end{equation}
where $\alpha = 0.05$ is used to target 95\% marginal coverage. The resulting quantile vector $q \in \mathbb{R}^{n_e}$ is reshaped to match the output grid.

\begin{figure}[!htbp]
    \centering
    \includegraphics[width=1.00\textwidth]{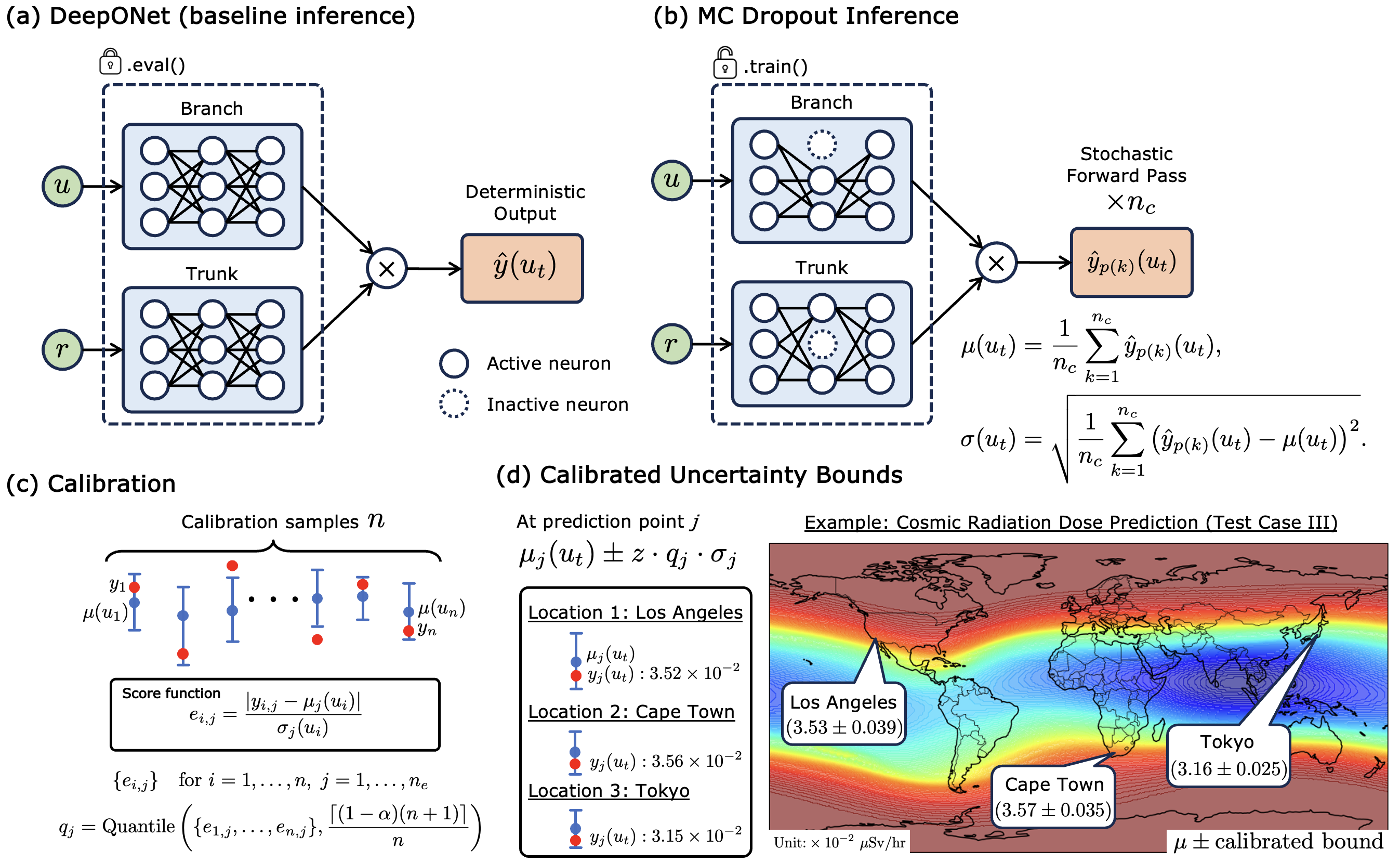}
    \caption{Overview of the proposed conformal UQ framework. 
    (a) Baseline DeepONet inference with dropout off (deterministic output). 
    (b) MC dropout inference enables multiple stochastic predictions. 
    (c) Calibration using normalized residuals yields location-specific quantiles. 
    (d) Final prediction intervals are constructed as 
    $\mu_j(u_t) \pm z \cdot q_j \cdot \sigma_j(u_t)$. 
    Example shown: cosmic radiation dose prediction from \textbf{Test Case 3}. 
    \textbf{Unit:}~$\times 10^{-2}~\mu\mathrm{Sv/hr}$.
}
    \label{fig:framework}
\end{figure}

The final calibrated prediction interval for each test input $u_t$ is then given by
\begin{equation}
\mathcal{I}(u_t) = \left[ \mu(u_t) - z \cdot q \cdot \sigma(u_t), \quad \mu(u_t) + z \cdot q \cdot \sigma(u_t) \right],
\end{equation}
where $z = 1.96$ denotes the Gaussian multiplier, and all operations are performed elementwise. Figure \ref{fig:framework} (c) represents the calibration task, and Figure \ref{fig:framework} (d) shows the example of final prediction with uncertainty bounds. 

To evaluate the effectiveness of the calibrated prediction intervals, three diagnostic metrics are used: empirical coverage, relative error, and failure rate.

The empirical coverage for the $i$-th test sample is defined as
\begin{equation}
C^{(i)} = \frac{1}{n_e} \sum_{j=1}^{n_e} \mathds{1} \left[ y_{i,j} \in \mathcal{I}_j(u_i) \right],
\label{eq:coverage}
\end{equation}
where $\mathds{1}[\cdot]$ denotes the indicator function, and $\mathcal{I}_j(u_i)$ is the calibrated prediction interval at spatial index $j$. This metric quantifies the fraction of output locations for which the predicted confidence band includes the true value.

To assess point prediction accuracy, the mean relative error for the $i$-th test sample is computed as
\begin{equation}
\epsilon^{(i)} = \frac{1}{n_e} \sum_{j=1}^{n_e} \left| \frac{ \mu_j(u_i) - y_{i,j} }{ y_{i,j} + \varepsilon } \right| \times 100\%,
\end{equation}
where $\mu_j(u_i)$ is the ensemble mean prediction at location $j$, and $\varepsilon$ is a small constant added for numerical stability.

To quantify the extent of local under-coverage, the failure rate for the $i$-th test sample is defined as
\begin{equation}
F^{(i)} = \frac{1}{n_e} \sum_{j=1}^{n_e} \mathds{1} \left[ y_{i,j} \notin \mathcal{I}_j(u_i) \right] \times 100\%,
\end{equation}
which measures the proportion of output locations where the calibrated interval fails to capture the ground truth value.

Test samples exhibiting both high failure rate and high relative error are flagged as critical outliers, as they represent scenarios in which the model is simultaneously inaccurate and overconfident. These cases are of particular concern from a reliability standpoint and are visualized to assess the spatial distribution and severity of predictive failures.

\section{Description of test cases}

\subsection{Lid-Driven Cavity}
%

\begin{figure}[!htbp]
    \centering
    \includegraphics[width=0.75\textwidth]{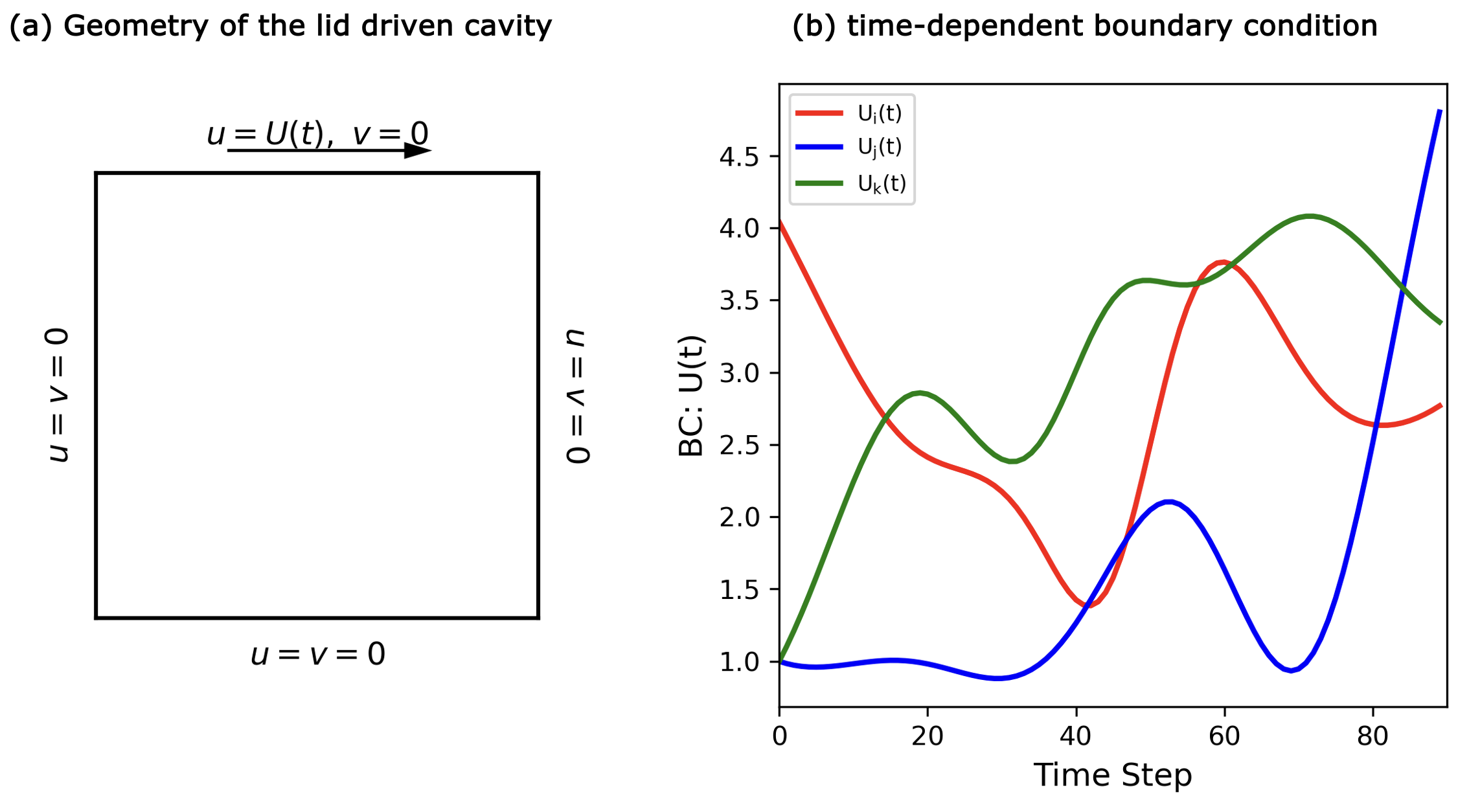}
    \caption{Lid-driven cavity setup with time-dependent lid velocity $ U(t) $. (a) Domain geometry and boundary conditions. (b) Example input functions $U(t)$ used in simulations. Outputs include velocity magnitude, pressure, and TKE on a 422 node mesh.}
    \label{fig:ldc_setup}
\end{figure}

The two-dimensional incompressible lid-driven cavity (LDC) flow is considered in a unit square domain $\Omega = [0,1] \times [0,1]$ as shown in Figure \ref{fig:ldc_setup} (a). The flow is induced by a time-dependent horizontal velocity applied to the top boundary, while the remaining walls are kept stationary under no-slip conditions. 

The governing equations are incompressible Navier-Stokes equations formulated under the Reynolds-averaged Navier-Stokes (RANS) model. The velocity field $\mathbf{u}=(u,v)$ and pressure field $p$ satisfy the continuity condition $\nabla \cdot \mathbf{u}=0$ and the momentum conservation equation:

\begin{equation}
\frac{\partial \mathbf{u}}{\partial t} + (\mathbf{u} \cdot \nabla)\mathbf{u} = -\nabla p + \nabla \cdot \left[ \left( \nu + \nu_t \right) \left( \nabla \mathbf{u} + \nabla \mathbf{u}^\top \right) \right]
\end{equation}

where $\nu$ denotes the kinematic viscosity, and $v_t$ is the eddy viscosity provided by a turbulence closure model. 

The effects of turbulence are represented through the turbulent kinetic energy (TKE) $k(x,y,t)$, which is governed by the transport equation:

\begin{equation}
\frac{\partial k}{\partial t} + \mathbf{u} \cdot \nabla k = P_k - \varepsilon + \nabla \cdot \left[ \left( \nu + \frac{\nu_t}{\sigma_k} \right) \nabla k \right]
\end{equation}

where $P_k$ represents the production of TKE, $\epsilon$ is the dissipation rate, and $\sigma_k$ is a model parameter. 

A time-dependent boundary condition of the form $u(x,y=1, t)=U(t),v=0$ is applied along the top lid, while no-slip conditions ($u=v=0$) are enforced on the bottom and side walls. The fluid is initialized at rest with $\mathbf{u}(x,y,0)=(0,0)$, and a uniform pressure field $p_0$. Several representative input profiles for the lid velocity $U(t)$ are shown in Figure \ref{fig:ldc_setup} (b). Each input is discretized over 90 time steps to express the temporal variation of the boundary forcing, yielding a one-dimensional input of shape of [90, 1]. Simulations are conducted on a unstructured mesh comprising 4225 spatial nodes with ANSYS Fluent. At each time step, the solver outputs the velocity magnitudes $|\mathbf{u}|=\sqrt{u^2 + v^2}$, pressure field $p(x,y,t)$, and TKE $k(x,y,t)$. In this work, only the TKE field is selected as the target variable for modeling.

The goal is to learn an operator \( \mathcal{G}_\theta \) that maps the lid velocity history \( U(t) \in \mathbb{R}^{90 \times 1} \) to the spatial distribution of TKE at the final simulation time step:

\begin{equation}
\mathcal{G}_\theta : U(t) \mapsto k(x, y, t=1) \in \mathbb{R}^{4225}.
\end{equation}

The DeepONet architecture is employed; the branch network processes the temporal input function $ U(t) $, which governs the evolution of the lid velocity. A four-layer GRU with 256 hidden units is used to extract a latent representation of the input sequence. The trunk network processes the spatial coordinates $(x, y) \in \Omega $ at which the solution is evaluated. It consists of a fully connected neural network with four hidden layers, each comprising 128 to 256 neurons, hyperbolic tangent (\texttt{Tanh}) activations, and dropout with a rate of $ p = 0.1 $. The final prediction of TKE at each spatial node is obtained via an inner product between the outputs of the branch and trunk networks.

Training is performed using the Adam optimizer with a learning rate of \( 10^{-4} \), and the learning rate is adaptively adjusted using a plateau-based decay scheduler. The model is trained for 200 epochs with a batch size of 64, and the mean squared error (MSE) is employed as a loss function.


\subsection{Plastic deformation}
This case study addresses the modeling of history-dependent plastic deformation in a two-dimensional domain under small-strain assumptions. The governing equations are expressed in terms of the Cauchy stress tensor \(\boldsymbol{\sigma}\), where mechanical equilibrium is enforced throughout the spatial domain \(\Omega\) via the balance law:

\begin{equation}
    \nabla \cdot \boldsymbol{\sigma} = \mathbf{0}, \quad \forall \vec{x} \in \Omega,
\end{equation}

together with the Dirichlet and Neumann boundary conditions:

\begin{equation}
    \mathbf{u} = \bar{\mathbf{u}}, \quad \forall \vec{x} \in \partial \Omega_u,
\end{equation}
\begin{equation}
    \boldsymbol{\sigma} \cdot \mathbf{n} = \bar{\mathbf{t}}, \quad \forall \vec{x} \in \partial \Omega_t,
\end{equation}

where \(\mathbf{u}\) is the displacement field, \(\bar{\mathbf{u}}\) is the prescribed boundary displacement, \(\bar{\mathbf{t}}\) is the prescribed traction, and \(\mathbf{n}\) is the outward unit normal vector on the boundary.

Under the small deformation assumption, the total strain tensor \(\boldsymbol{\varepsilon}\) is defined as:

\begin{equation}
    \boldsymbol{\varepsilon} = \frac{1}{2} \left( \nabla \mathbf{u} + \nabla \mathbf{u}^\top \right)
\end{equation}

The strain is decomposed into elastic and plastic parts:

\begin{equation}
    \boldsymbol{\varepsilon} = \boldsymbol{\varepsilon}^e + \boldsymbol{\varepsilon}^p
\end{equation}

For a linear elastic, isotropic material under plane stress conditions, the constitutive relation is given by:

\begin{equation}
    \begin{bmatrix}
    \sigma_{11} \\
    \sigma_{22} \\
    \sigma_{12}
    \end{bmatrix}
    =
    \frac{E}{1 - \nu^2}
    \begin{bmatrix}
    1 & \nu & 0 \\
    \nu & 1 & 0 \\
    0 & 0 & \frac{1 - \nu}{2}
    \end{bmatrix}
    \begin{bmatrix}
    \varepsilon_{11} \\
    \varepsilon_{22} \\
    \varepsilon_{12}
    \end{bmatrix}
\end{equation}

where \(E\) is Young’s modulus and \(\nu\) is Poisson’s ratio.

Plasticity is modeled using \(J_2\) flow theory with linear isotropic hardening. The yield stress evolves with the accumulated equivalent plastic strain \(\bar{\varepsilon}_p\) as:

\begin{equation}
    \sigma_y(\bar{\varepsilon}_p) = \sigma_{y0} + H \bar{\varepsilon}_p
\end{equation}

where \(\sigma_y\) is the current yield stress, \(\sigma_{y0}\) is the initial yield stress, and \(H\) is the hardening modulus. This formulation captures the path-dependent stress response of the material under time-varying boundary loading and forms the basis for learning a surrogate model that maps loading histories to the final stress field.

\begin{figure}[!htbp]
    \centering
    \includegraphics[width=0.75\textwidth]{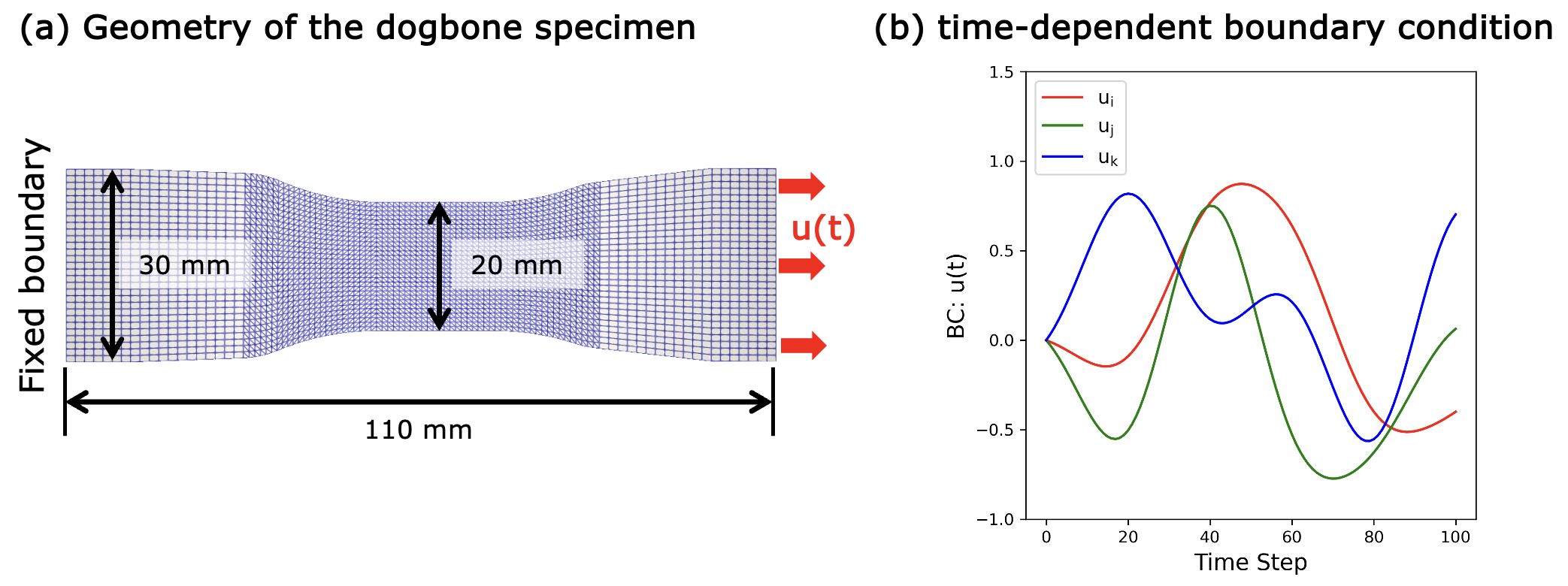}
    \caption{(a) Geometry and mesh of the dogbone specimen with fixed left boundary and time-dependent displacement $u(t)$ applied on the right. (b) Examples of time-varying displacement boundary conditions used as model inputs.}
    \label{fig:plastic_setup}
\end{figure}

The simulation dataset used in this case study was generated using a high-fidelity finite element model of a dog-bone-shaped specimen subjected to time-varying axial displacement shown in Figure \ref{fig:plastic_setup} (a). The geometry consists of a 110 mm long planar bar with a reduced gauge width of 20 mm and wider grip sections of 30 mm. The specimen is discretized into 4756 linear plane stress elements composed of a combination of four-node quadrilateral and three-node triangular elements, with a uniform thickness of 1 mm.

The boundary conditions reflect a standard tension test: the left edge of the specimen is fully constrained, and a displacement is applied along the global $x$-direction on the right edge. The displacement profile is time-dependent and defined over the interval $t \in [0, 1]$ seconds as shown in Figure \ref{fig:plastic_setup} (b). Each loading history is parameterized by six displacement control points—two fixed at the start and end times (0 and 1 s), and four additional points sampled randomly from the interval $[0.1, 0.9]$ s. These values are then interpolated using radial basis functions to produce a smooth, continuously differentiable displacement curve. The displacement magnitudes are selected to ensure that the maximum nominal axial strain remains below 5\%, maintaining consistency with the small-strain formulation.

A total of 15{,}000 simulations were performed using Abaqus/Standard with implicit integration. Each simulation corresponds to a unique loading history, and the result of interest is the von Mises stress field $\bar{\sigma}(\vec{r})$ at the final time step. The stress field is defined over the 2D spatial domain and computed from the in-plane Cauchy stress components using the expression:

\begin{equation}
    \bar{\sigma} = \sqrt{\sigma_{11}^2 + \sigma_{22}^2 - \sigma_{11} \sigma_{22} + 3\sigma_{12}^2}
\end{equation}

This dataset, consisting of input displacement time histories and corresponding output stress fields, is used to train a surrogate model that learns the mapping from time-varying boundary loads to spatially distributed stress responses in a history-dependent, path-sensitive material system.

A DeepONet architecture is employed to learn the mapping from time-varying displacement histories to the final von Mises stress field. The operator learning task is defined as $ \mathcal{G}_\theta: u(t) \mapsto \bar{\sigma}(x, y) $, where $ u(t) \in \mathbb{R}^{101 \times 1} $ represents the discretized displacement profile over the interval $ t \in [0,1] $, and $ \bar{\sigma}(x, y) \in \mathbb{R}^{4756} $ denotes the spatially distributed stress field at the final time step.

The branch network is implemented using a GRU with four layers and 256 hidden units per layer. The trunk network is constructed as a fully connected neural network comprising four layers with 128 to 256 neurons, hyperbolic tangent (\texttt{Tanh}) activation functions, and a dropout rate of $ p = 0.1 $. The final stress prediction at each spatial location is computed, producing a scalar output at each of the 4756 spatial nodes.

Model training is performed using the Adam optimizer with a base learning rate of $ 10^{-4} $, and learning rate scheduling is applied via a plateau-based decay method. The model is trained for up to 200 epochs with mini-batches of size 64.

\subsection{Cosmic radiation dose}
Monitoring effective dose rates from cosmic radiation is vital for public radiation protection, aviation safety, and space weather forecasting. While radiation exposure in aviation and orbital environments has been extensively characterized, the terrestrial dose from galactic cosmic rays (GCRs) and solar energetic particles (SEPs) remains difficult to predict, especially during solar storms and geomagnetic disturbances. Accurate estimation of ground-level dose is also critical for designing radiation-hardened aerospace systems and supporting rare-event physics experiments in underground laboratories and neutrino observatories.

Ground-based neutron monitors provide continuous observations but are limited in scope: they record only secondary neutrons and do not capture other dose-contributing components such as muons, electrons, or photons. Moreover, effective dose is a biologically weighted function of particle type and energy spectrum, making it difficult to infer from neutron counts alone. High-fidelity Monte Carlo simulations, such as those based on GEANT4 or PHITS, can model atmospheric propagation and dose deposition with precision, but are computationally prohibitive for large-scale or real-time applications. To over come those issues, the authors' prior work \cite{kobayashi2025proxies} proposed the model maps from sparse neutron monitor observations to effective dose estimates across multiple locations. In this work, we embed UQ into the model to identify epistemic risk, detect anomalous space weather events, and ensure safety in downstream applications. It enables reliable and interpretable predictions in safety-critical domains where overconfidence or underestimation can lead to severe consequences.

The effective dose rate at sea level due to cosmic radiation was prepared using the PHITS-based Analytical Radiation Model in the Atmosphere (PARMA) \cite{sato2015analytical,sato2016analytical,expacs}. PARMA analytically parameterizes extensive air shower simulations performed using the PHITS Monte Carlo transport code \cite{iwamoto2022benchmark,sato2024recent}, modeling the atmospheric interactions of galactic cosmic rays and the resulting secondary particle fluxes. The simulation domain is defined as a discrete spatial grid \(\Omega \subset \mathbb{R}^2\), where each spatial location is represented by a vector \(\vec{r} = (x, y)\), with \(x\) and \(y\) corresponding to longitude and latitude, respectively. The domain is uniformly discretized at a resolution of \(1^\circ \times 1^\circ\), resulting in 65{,}341 spatial points across the Earth’s surface. For each location \(\vec{r} \in \Omega\), the effective dose rate was computed daily over a 23-year period from January 1, 2001, to December 31, 2023, yielding 8400 time steps and more than 550 million spatiotemporal data points.

At each time step, the simulation evaluates the effective dose rate $D(\vec{r},t)$ by integrating energy-dependent secondary particle fluxes over all major species:

\begin{equation}
    D(\vec{r}, t) = \sum_{i \in \mathcal{P}} \int_0^\infty \phi_i(E; \vec{r}, t) \cdot w_i(E) \, dE
\end{equation}

where \(\mathcal{P} = \{n, \mu^{\pm}, e^{\pm}, \gamma\}\) denotes the set of particle types contributing to the dose, \(\phi_{i}(E; \vec{r}, t)\) is the differential energy spectrum of species \(i\), and \(w_{i}(E)\) is the corresponding fluence-to-dose conversion coefficient based on ICRP recommendations.

The complete simulation dataset can therefore be represented as a two-dimensional array \(\mathbf{D} \in \mathbb{R}^{8400 \times 65341}\), where each row corresponds to a daily time step and each column to a spatial location \(\vec{r} \in \Omega\).
To develop a surrogate model for predicting \(D(\vec{r}, t)\), we use as inputs the spatial location \(\vec{r}\) and time-dependent neutron monitor observations. The neutron monitor data used in this study were obtained from twelve globally distributed stations over the same 8400-day period as the simulation. For each day \(t\), the average neutron count rates (in counts/hour) from these stations are concatenated into a feature vector \(\mathbf{n}(t) \in \mathbb{R}^{12}\), representing the global neutron activity. Collectively, these vectors form a matrix \(\mathbf{N} \in \mathbb{R}^{8400 \times 12}\), where each row corresponds to a specific day and each column to a specific neutron monitor station.

To incorporate short-term temporal dynamics into the surrogate model, both the dose dataset \(\mathbf{D}\) and neutron monitor dataset \(\mathbf{N}\) are sequentialized using a fixed sequence length of 7 days. For each prediction at time \(t\), the model is conditioned not only on the current neutron data \(\mathbf{n}(t)\), but also on the preceding six days, forming a 7-day temporal window. The resulting input is a matrix \(\mathbf{N}_{t}^{(7)} \in \mathbb{R}^{7 \times 12}\), representing the neutron monitor readings from day \(t-6\) to \(t\), while the associated target remains the effective dose rate \(D(\vec{r}, t)\) at the final time step.

The neural operator trained in this problem learns a mapping from a sequence of time-dependent inputs to a spatially distributed scalar field. Specifically, the operator \(\mathcal{G}_\theta\), parameterized by trainable parameters \(\theta\), takes the 7-day neutron monitor sequence \(\mathbf{N}_t^{(7)}\) as input and outputs a function defined over spatial coordinates \(\vec{r} \in \Omega\). The predicted effective dose rate is expressed as:
\begin{equation}
    \hat{D}(\vec{r}, t) = \mathcal{G}_\theta(\mathbf{N}_t^{(7)})(\vec{r})
\end{equation}

For this case, a Sequential DeepONet architecture was employed to learn the mapping from time-series neutron monitor readings to the global spatial distribution of cosmic radiation dose. The branch network was implemented as a four-layer GRU with 256 hidden units, designed to encode 7-day sequences of 12-dimensional neutron monitor features. Like the plastic deformation case, Layer normalization was applied to the GRU output. The trunk network processed 2D spatial coordinates through a fully connected network consisting of four layers with 256 neurons, ReLU activations, and dropout (with a rate of $p = 0.1$). The final effective dose prediction at each spatial location was obtained via an inner product between the trunk and branch outputs. The model comprised approximately 1.45 million trainable parameters.

\section{Results}
Deterministic models were trained for each case study and evaluated on hold-out test datasets to evaluate the predictive performance of the neural operator models prior to uncertainty calibration. Table~\ref{tab:dataset-sizes} summarizes the dataset sizes used for training, calibration, and testing.

\begin{table}[htpb]
\caption{Dataset size used in case studies.}
\centering
\label{tab:dataset-sizes}
\resizebox{0.5\textwidth}{!}{%
\begin{tabular}{@{}lccc@{}}
\toprule
\multirow{2}{*}{Test Cases} & \multicolumn{3}{l}{No. of Samples in Dataset} \\ \cmidrule(l){2-4} 
 & Training & Calibration & Test \\ \midrule
Case I: Lid-Driven Cavity & 3949 & 493 & 495 \\
Case II: Plastic Deformation & 11920 & 2384 & 100 \\
Case III: Cosmic Radiation Dose & 6422 & 1601 & 359 \\ \bottomrule
\end{tabular}%
}
\end{table}

The model evaluation metrics, including mean absolute error (MAE), root mean square error (RMSE), relative $L_2$ error, and peak signal-to-noise ratio (PSNR, where applicable), were used for performance evaluation. The list of evaluation metrics is summarized in Table \ref{tab:model_performance}. All results are based on the point predictions of a single trained model for each case, without ensembling or dropout.

In Case I (Lid-Driven Cavity), the model achieved a relative $ L_2 $ error of 4.91\%, which is the highest among the three cases. This elevated error may be attributed to the complexity of turbulence-like flow structures in the cavity, which are characterized by sharp gradients and spatial variability that challenge surrogate model generalization. 

In Case II (Plastic Deformation), the model achieved a relative $L_2$ error of 3.56\%, indicating improved accuracy compared to Case I. It is worth noting that a prior study by He et al. \cite{he2024sequential} reported a mean relative $L_2$ error of 0.847\% (with a standard deviation of 2.853\%) for the same problem using a GRU-based DeepONet. In our study, the reported error is slightly higher. This discrepancy may be attributed to differences in implementation details, including data normalization schemes, batch size, training dataset size, and model parameter settings. 

In Case III (Cosmic Radiation Dose), the model achieved the lowest errors across all metrics. The relative $L_2$ error was approximately 3.94\%, and the PSNR reached 49.30~dB. This performance likely results from the smoothness and spatiotemporal regularity of the dose field, which makes the mapping more amenable to operator learning from sparse proxy inputs.

\begin{table}[htbp]
\centering
\caption{Test performance of the base models for each case.}
\label{tab:model_performance}
\resizebox{0.75\textwidth}{!}{%
\begin{tabular}{@{}lcccc@{}}
\toprule
Test Cases & MAE & RMSE  & Relative L2 error (\%) & PSNR (dB) \\ \midrule
Case I: Lid-Driven Cavity & $4.23\times 10^{-3}$ & $6.71 \times 10^{-3}$ & 4.91 & - \\
Case II: Plastic Deformation & 1.92 & 3.13  & 3.56 & - \\
Case III: Cosmic Radiation Dose & $1.43 \times 10^{-4}$ & $1.54 \times 10^{-4}$ & $3.94 \times 10^{-2}$ & 49.30 \\ \bottomrule
\end{tabular}%
}
\end{table}

For each base model listed in Table \ref{tab:model_performance}, 10 ensemble predictions were generated by activating Monte Carlo dropout at inference time. These samples were used to perform conformal calibration, resulting in 95\% prediction intervals for each test case. The empirical coverage of the calibrated intervals was then evaluated using the metric defined in Equation \eqref{eq:coverage}, which computes the fraction of spatial locations where the prediction interval includes the true value. Table~\ref{tab:coverage_analysis} summarizes the resulting coverage statistics, including the average, minimum, and maximum coverage values across the test set and the count of samples that achieved coverage above or below the nominal 95\% level.

Case I yielded an average coverage of 99.68\%, with a minimum of 92.59\%, and only 4 out of 495 samples falling below the target level. Case II achieved an average coverage of 98.96\%, with a minimum of 53.33\%, and 4 out of 100 samples undercovered. In Case III, the average coverage was 98.63\%, while the minimum dropped to 1.14\%, with 15 out of 359 samples showing under coverage. Although a few test cases presented local failures, the method maintained valid marginal coverage across most spatial queries, consistent with the theoretical guarantees of conformal prediction. A more detailed analysis of prediction errors and uncertainty behavior for each test case is presented in the following subsections.

\begin{table}[htbp]
\centering
\caption{Coverage analysis of the 95\% prediction intervals for each test case.}
\begin{tabular}{@{}lccccc@{}}
\toprule
\multirow{2}{*}{Test Case} & \multicolumn{3}{c}{Coverage (\%)} & \multicolumn{2}{c}{Sample Count} \\ 
\cmidrule(lr){2-4} \cmidrule(l){5-6}
 & Average & Min & Max & $\geq$ 95\% & $< 95\%$ \\ \midrule
Case I: Lid-Driven Cavity     &  99.68    & 92.59    & 100.00      & 491   & 4 \\
Case II: Plastic Deformation   & 98.96  & 53.33 & 100.00  & 96   & 4   \\
Case III: Cosmic Radiation Dose & 98.63  & 1.14  & 100.00  & 344  & 15  \\ 
\bottomrule
\label{tab:coverage_analysis}
\end{tabular}
\end{table}

\subsection{Case I: Lid-driven cavity}
\begin{figure}[!htbp]
    \centering
    \includegraphics[width=0.75\textwidth]{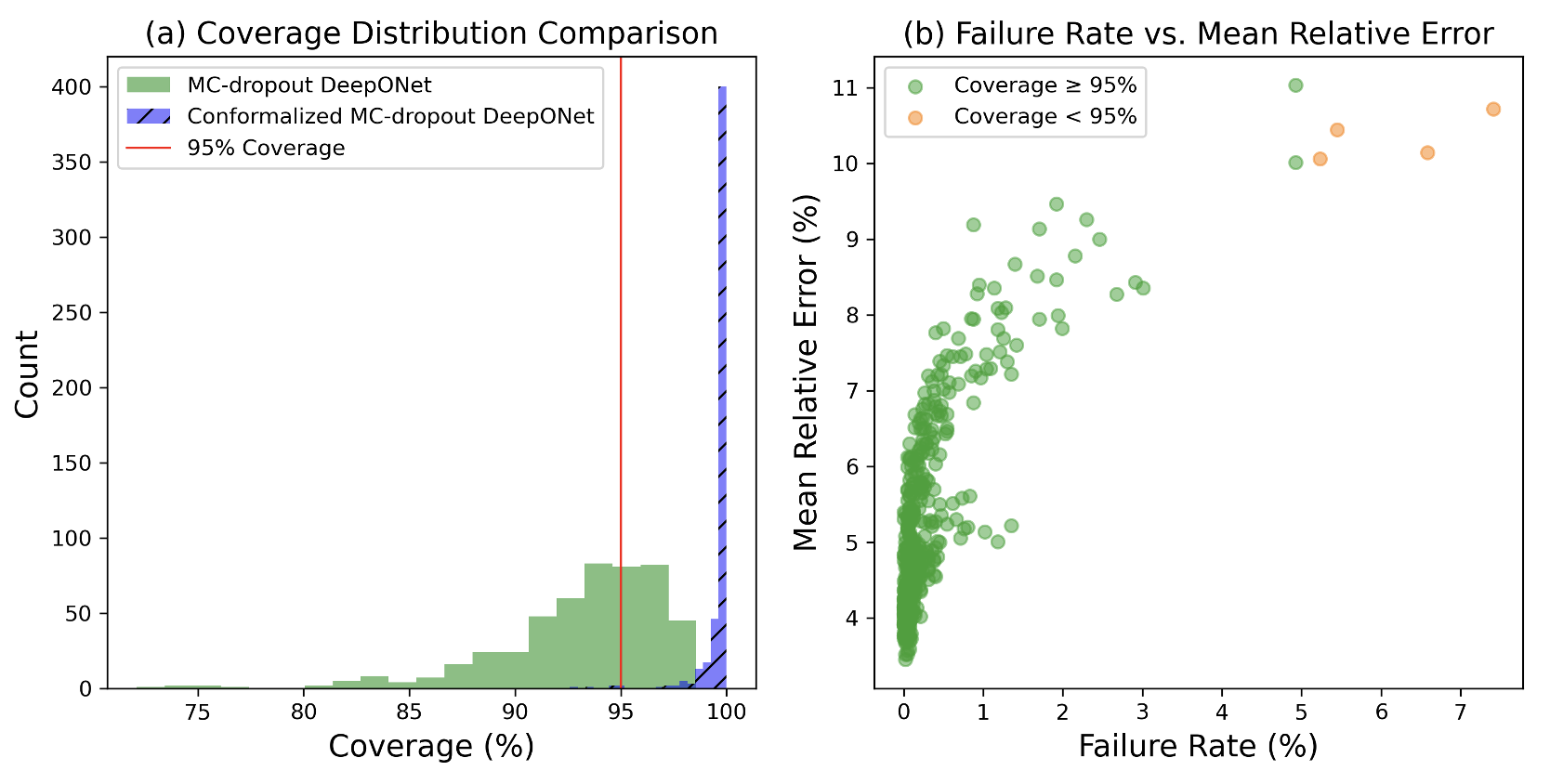}
    \caption{Uncertainty quantification analysis of Conformalized MC-dropout DeepONet.
(a) Empirical coverage distribution across test samples.
(b) Scatter plot of failure rate vs. mean relative error. Samples are color-coded based on whether their coverage is above or below the nominal 95\% threshold.}
    \label{fig:ldc_analysis}
\end{figure}

Figure \ref{fig:ldc_analysis} quantitatively evaluates uncertainty calibration performance for the lid-driven cavity case using the Conformalized MC-dropout DeepONet. Panel (a) shows the empirical coverage distribution across all test samples. The histogram highlights a key distinction between the uncalibrated MC-dropout model and the conformalized version. While the raw MC-dropout intervals tend to undercover, the calibrated intervals consistently align around the nominal 95\% level, demonstrating the effectiveness of conformal correction in achieving statistically valid uncertainty estimates.

Panel (b) presents a scatter plot of failure rate versus mean relative error for each test sample. Samples are color-coded based on whether their empirical coverage is above or below the 95\% target. A clear trend emerges: samples with coverage below 95\% (orange) exhibit higher prediction errors, whereas those meeting or exceeding the threshold (green) generally achieve lower mean errors. It suggests a correlation between model confidence and prediction quality and further supports the utility of failure rate analysis as a diagnostic tool.

\begin{figure}[!htbp]
    \centering
    \includegraphics[width=\textwidth]{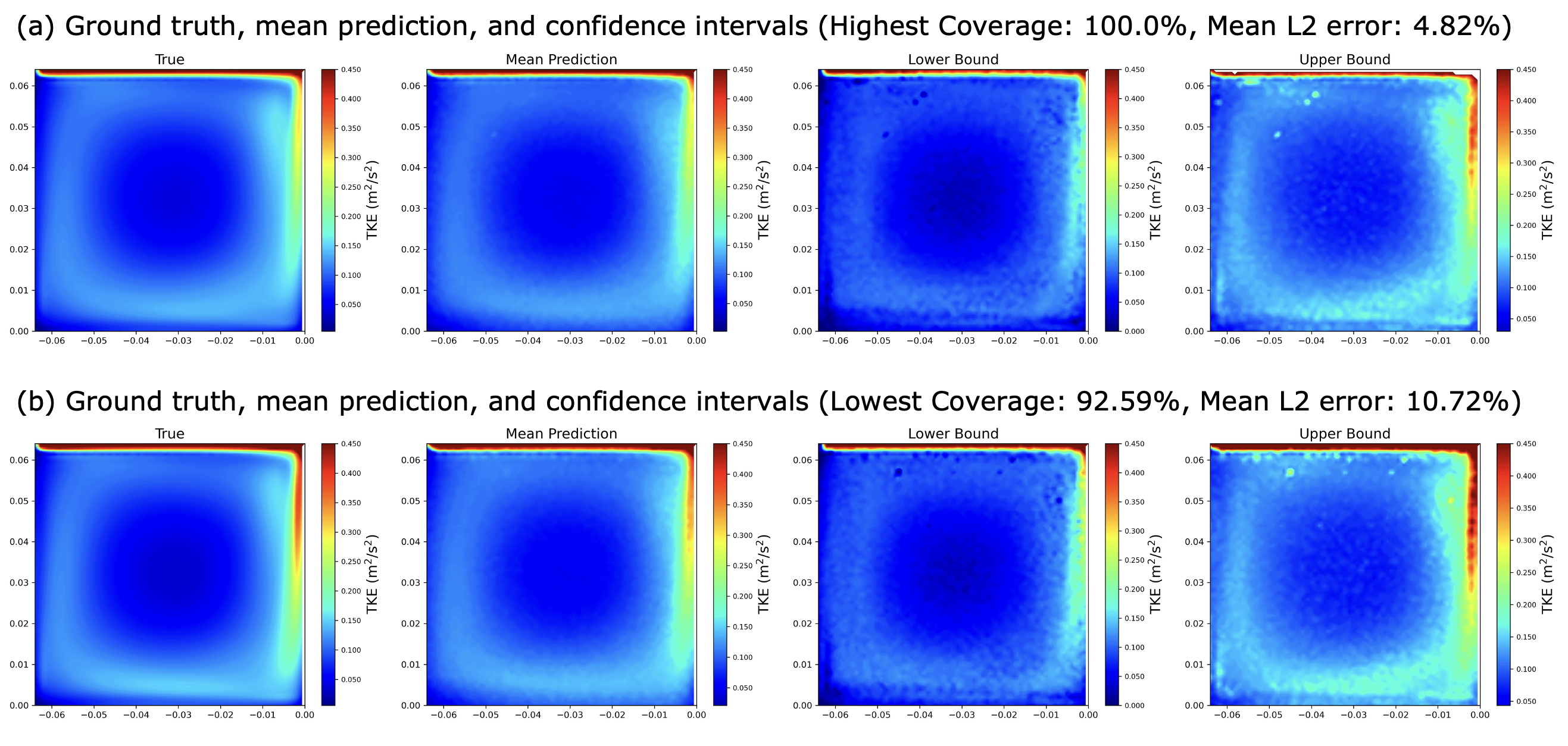}
    \caption{Ground truth, mean prediction, and calibrated confidence intervals of turbulence kinetic energy (TKE) for the lid-driven cavity problem. (a) Sample with highest empirical coverage (100\%) and lowest mean relative L2 error (4.82\%). (b) Sample with lowest empirical coverage (92.59\%) and higher error (10.72\%).}
    \label{fig:ldc_plot}
\end{figure}

Figure \ref{fig:ldc_plot} visualizes two representative samples from the lid-driven cavity test set. The top row (\ref{fig:ldc_plot} (a)) corresponds to the best-performing case, achieving 100\% empirical coverage and a low relative $L_2$ error of 4.82\%. The predicted mean and confidence bounds closely match the ground truth, particularly in regions with strong shear near the upper wall. In contrast, the bottom row (\ref{fig:ldc_plot} (b)) shows the worst-covered sample, with 92.59\% coverage and a higher relative error of 10.72\%. The prediction intervals are visibly looser and miss high-gradient regions near the cavity corners, where undercover and model bias tend to co-occur.

Together, these results highlight the conformalized MC-dropout DeepONet's reliability and limitations. While most predictions are well-calibrated, localized under coverage can still occur in regions with high spatial complexity or unresolved dynamics.

\subsection{Case II: Plastic deformation}
\begin{figure}[!htbp]
    \centering
    \includegraphics[width=0.75\textwidth]{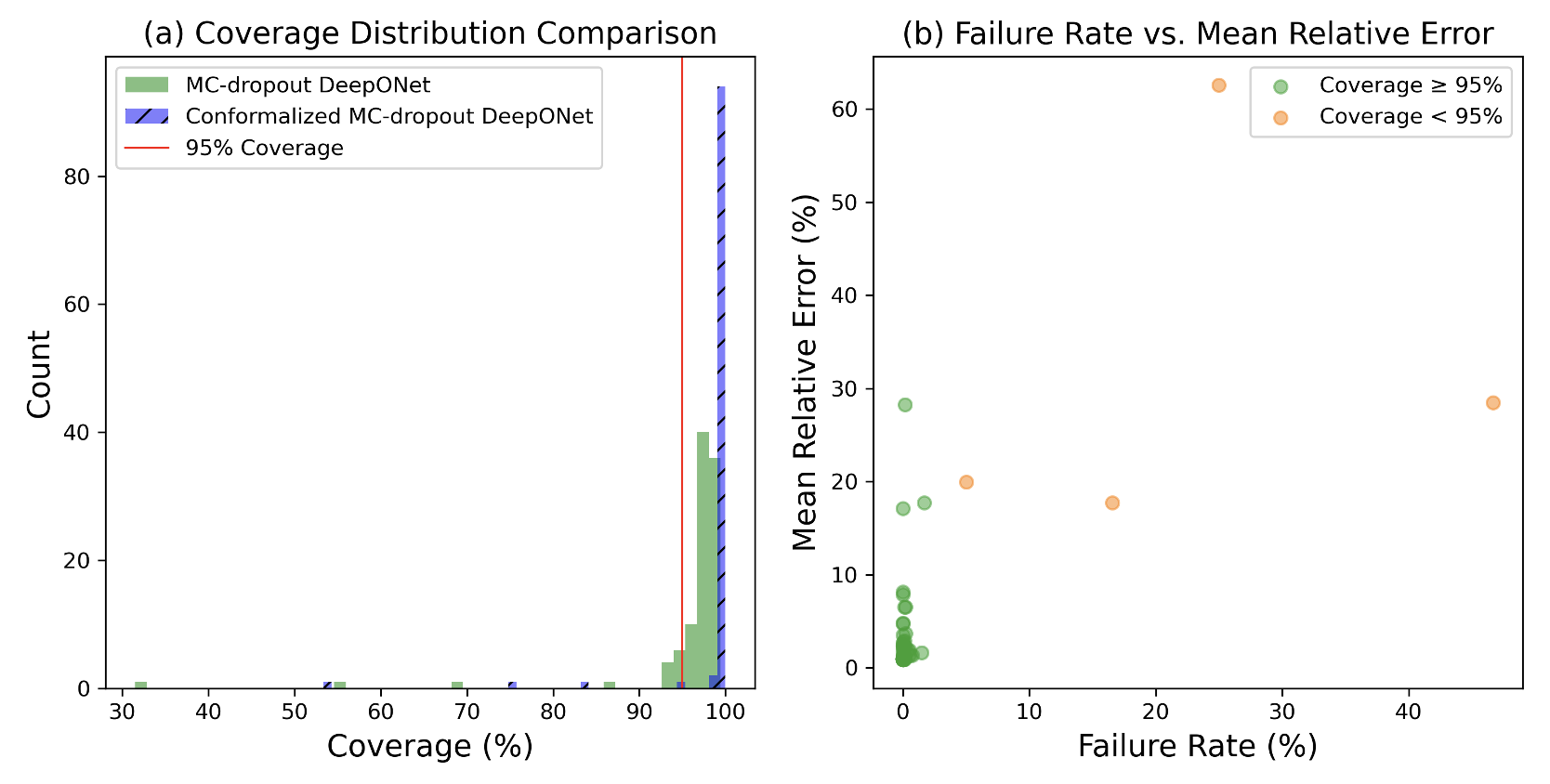}
    \caption{Uncertainty quantification analysis of Conformalized MC-dropout DeepONet.
(a) Empirical coverage distribution across test samples.
(b) Scatter plot of failure rate vs. mean relative error. Samples are color-coded based on whether their coverage is above or below the nominal 95\% threshold.}
    \label{fig:plastic_analysis}
\end{figure}

Following the previous analysis of the lid-driven cavity flow, Figure \ref{fig:plastic_analysis} presents the uncertainty quantification results for the plastic deformation test case using the Conformalized MC-dropout DeepONet. In panel (a), the empirical coverage distribution shows that conformal calibration successfully aligns predictions closer to the 95\% target. In contrast, the uncalibrated MC-dropout model exhibits higher variability and frequent under-coverage. Panel (b) further reveals that some test samples with low coverage also experience elevated mean relative errors. These underperforming cases, highlighted in orange, exhibit failure rates exceeding 10\% and errors above 20\%, suggesting that under-coverage is often associated with poor prediction quality.

\begin{figure}[!htbp]
    \centering
    \includegraphics[width=\textwidth]{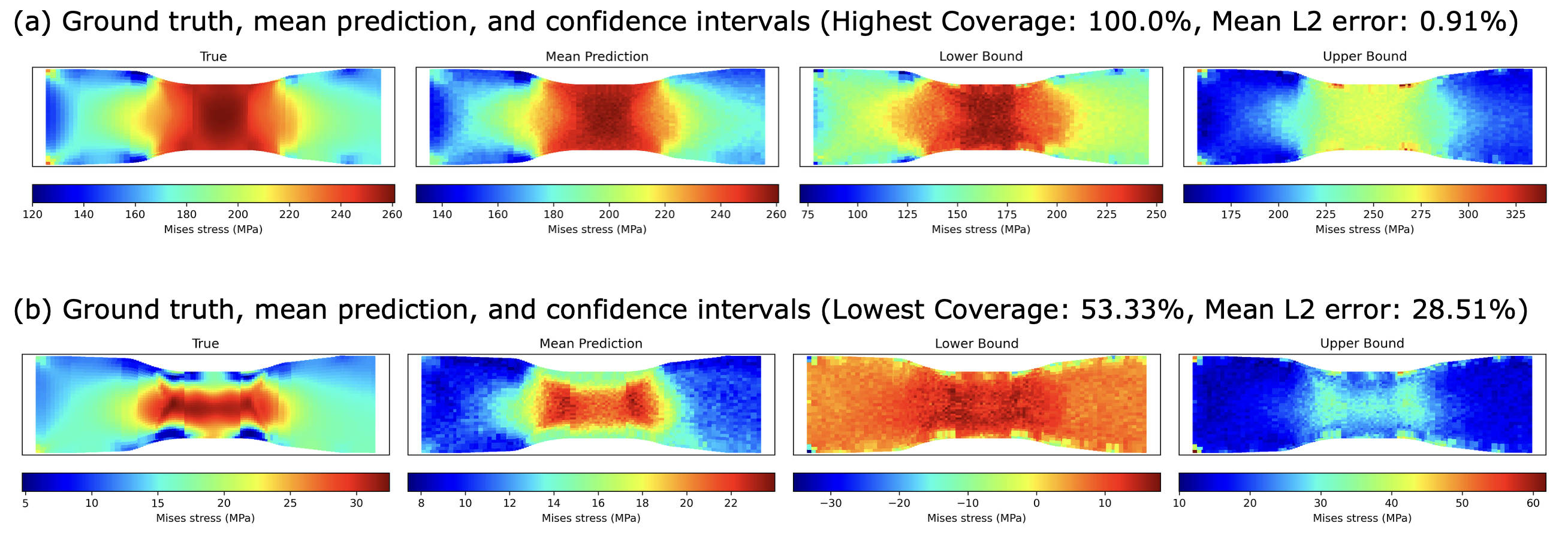}
    \caption{Ground truth, mean prediction, and calibrated confidence intervals for the plastic deformation task.
    (a) Sample with 100\% empirical coverage and low relative $L_2$ error (0.91\%), showing accurate predictions and well-calibrated uncertainty.
    (b) Sample with 53.33\% empirical coverage and high error (28.51\%), where the model underestimates uncertainty in regions with localized stress concentration.}
    \label{fig:plastic_plot}
\end{figure}

To examine this further, Figure \ref{fig:plastic_plot} compares two representative samples. The top row illustrates a sample with perfect empirical coverage of 100\% and a low relative $L_2$ error of 0.91\%. The predicted stress field agrees with the ground truth, and the prediction intervals appropriately capture the uncertainty, especially near regions with stress concentration. In contrast, the bottom row corresponds to the sample with the lowest coverage of 53.33\% and a high error of 28.51\%. In this case, the model underestimates uncertainty near the localized necking region, leading to substantial overconfidence and degraded predictive accuracy.

\subsection{Case III: Cosmic radiation dose}
\begin{figure}[!htbp]
    \centering
    \includegraphics[width=0.75\textwidth]{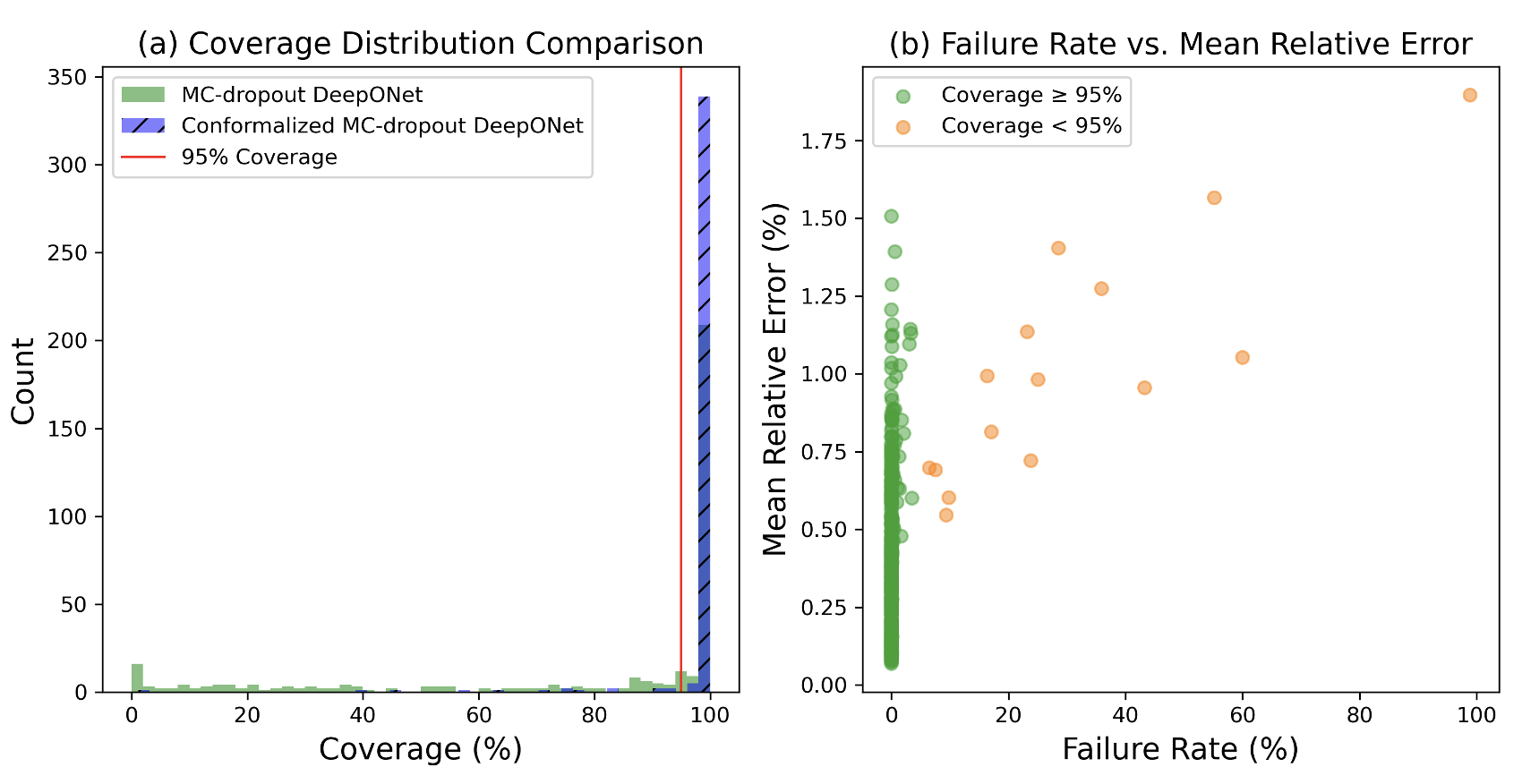}
    \caption{Uncertainty quantification analysis of Conformalized MC-dropout DeepONet.
(a) Empirical coverage distribution across test samples.
(b) Scatter plot of failure rate vs. mean relative error. Samples are color-coded based on whether their coverage is above or below the nominal 95\% threshold.}
    \label{fig:cosmic_analysis}
\end{figure}

Following the previous evaluations, Figure \ref{fig:cosmic_analysis} presents the uncertainty quantification performance for the cosmic radiation dose estimation task. As shown in panel (a), the conformalized MC-dropout DeepONet achieves consistently high empirical coverage, closely centered around the target 95\% level. In contrast, the uncalibrated MC-dropout model frequently undercovers, with many samples falling short of the nominal threshold. Panel (b) illustrates the relationship between failure rate and mean relative error across all test samples. Samples with coverage below 95\% (orange) tend to exhibit higher prediction errors, confirming that undercoverage correlates with reduced predictive accuracy even in globally smooth fields.

To qualitatively assess spatial behavior, Figure 9 compares two representative samples. The top row corresponds to the best-performing case, with 100\% empirical coverage and a low relative $L_2$ error of 0.31\%. The model accurately reconstructs the global dose rate distribution, and the prediction intervals are tightly aligned with the ground truth across latitudinal zones, including regions with high gradients, such as the South Atlantic Anomaly. In contrast, the bottom row shows the sample with the lowest empirical coverage of 1.24\% and a higher relative error of 1.90\%. Despite the smooth nature of the underlying field, the confidence intervals fail to adequately capture uncertainty near the equatorial and high-latitude regions, where sharp transitions or sensor sparsity may contribute to localized undercoverage.

Overall, the method performs robustly across most test samples, but the results indicate that conformal calibration may still underperform in specific geographic regions or under atypical solar conditions. It highlights the importance of spatial heterogeneity and real-world sensor limitations when calibrated UQ is applied to global-scale field estimation problems.

\begin{figure}[!htbp]
    \centering
    \includegraphics[width=\textwidth]{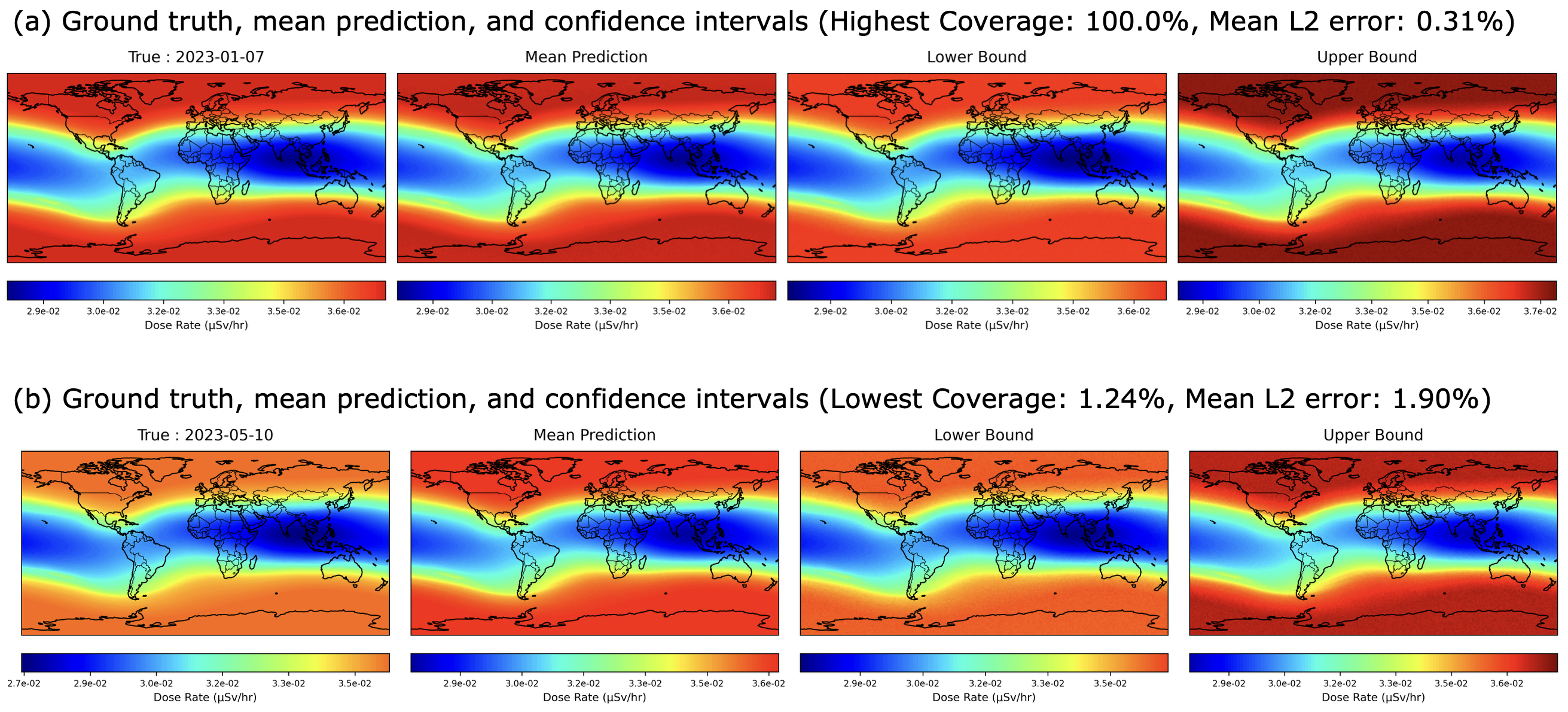}
    \caption{ (a) A well-calibrated case with high empirical coverage (100.0\%) and low mean relative L2 error (0.31\%), corresponding to January 7, 2023.
(b) A failure case with poor coverage (1.24\%) and higher mean relative L2 error (1.90\%), corresponding to May 10, 2023.
Each subplot displays the global spatial distribution of dose rates ($\rm{\mu Sv/hr}$), with consistent color scale limits for comparison. Prediction intervals were obtained using conformalized RP-DeepONet with distribution-free calibration.}
    \label{fig:cosmic_plot}
\end{figure}

\section{Conclusion}
This study introduced a calibrated uncertainty quantification framework for neural operator-based virtual sensing using Monte Carlo dropout and split conformal prediction. The proposed approach, the Conformalized Monte Carlo Operator (CMCO), enables the construction of distribution-free prediction intervals using a single-trained DeepONet model without requiring specialized loss functions or retraining. Combining dropout-based ensembling with conformal residual calibration produces spatially resolved confidence intervals that are both computationally efficient and theoretically guaranteed to achieve marginal coverage.

The framework was evaluated on increasing complexity in three physically grounded case studies: lid-driven cavity flow, elastoplastic deformation under time-dependent loading, and global cosmic radiation dose estimation from sparse neutron monitor inputs. Across all tasks, the method achieved high empirical coverage near the nominal 95\% level. Detailed analysis revealed that undercoverage was typically associated with high prediction errors and localized complexity, particularly in regions with strong gradients or sensor sparsity. Despite these challenges, the conformalized MC-dropout DeepONet maintained reliable uncertainty estimates across most test samples, demonstrating its practical utility for real-time virtual sensing.

Future work will explore adaptive calibration techniques that account for local data density and spatial heterogeneity. Extensions to multi-output fields, time-dependent targets, and broader classes of neural operators will also be considered to enhance generalizability and application breadth.

\section*{Replication of results}
The data and source code supporting this study will be available in the public GitHub repository \href{https://github.com/kkazuma19/}{https://github.com/kkazuma19/} upon the paper's acceptance.

\section*{Acknowledgments}

This research is a part of the Delta research computing project, which is supported by the National Science Foundation, (award OCI 2005572) and the State of Illinois, as well as the Illinois Computes program supported by the University of Illinois Urbana-Champaign and the University of Illinois System.

We acknowledge the NMDB database www.nmdb.eu, founded under the European Union's FP7 programme (contract no. 213007) for providing data. Additionally, we express our gratitude to the institutions and observatories that maintain and operate the individual neutron monitor stations, whose invaluable contributions made this work possible. Athens neutron monitor data were kindly provided by the Physics Department of the National and Kapodistrian University of Athens. Jungfraujoch neutron monitor data were made available by the Physikalisches Institut, University of Bern, Switzerland. Newark/Swarthmore, Fort Smith, Inuvik, Nain, and Thule neutron monitor data were obtained from the University of Delaware Department of Physics and Astronomy and the Bartol Research Institute. Kerguelen and Terre Adelie neutron monitor data were provided by Observatoire de Paris and the French Polar Institute (IPEV), France. Oulu and Dome C neutron monitor data were obtained from the Sodankylä Geophysical Observatory, University of Oulu, Finland, with support from the French-Italian Concordia Station (IPEV program n903 and PNRA Project LTCPAA PNRA14-00091). Apatity neutron monitor data were provided by the Polar Geophysical Institute of the Russian Academy of Sciences. South Pole neutron monitor data were supplied by the University of Wisconsin, River Falls.

\bibliographystyle{unsrt}  
\bibliography{references}  

\section*{Appendix}
\section*{Model Architecture for LDC}

The Sequential DeepONet architecture used for the lid-driven cavity problem consists of a recurrent branch network and a fully connected trunk network, which together model the mapping from the time-dependent lid velocity profile to the turbulent kinetic energy (TKE) field at the final simulation time. The detailed configuration is as follows:

\subsubsection*{Branch Network (Temporal Encoder)}
\begin{itemize}
    \item Type: LSTM
    \item Input size: 1 (scalar velocity at each time step)
    \item Number of layers: 4
    \item Hidden units per layer: 256
    \item Dropout: 0.1
    \item Output size: 256 (taken from final time step)
    \item Layer normalization applied to final hidden state
\end{itemize}

\subsubsection*{Trunk Network (Spatial Encoder)}
\begin{itemize}
    \item Type: Fully Connected Network (FCN)
    \item Architecture: [2, 512, 512, 512, 256]
    \item Activation function: ReLU
    \item Dropout rate: 0.1 (applied after each hidden layer)
    \item Input: 2D spatial coordinates \( (x, y) \)
    \item Output: 256-dimensional latent representation per spatial point
\end{itemize}

\subsubsection*{Output Construction}
The final output at each spatial location is computed via an inner product between the latent representations from the branch and trunk networks:

\[
\hat{k}(x_j, y_j) = \sum_{\ell=1}^{256} \phi_\ell(x_j, y_j) \cdot \psi_\ell(U(t)),
\]

where \( \psi \in \mathbb{R}^{256} \) is the encoded branch vector, and \( \phi(x_j, y_j) \in \mathbb{R}^{256} \) is the coordinate-dependent output of the trunk network. A learnable bias term is added per output channel.

\subsubsection*{Model Summary}
\begin{itemize}
    \item Total number of trainable parameters: \textbf{2,502,913}
    \item Number of outputs: 1 (scalar TKE value per spatial point)
    \item Batch processing: Branch features are broadcast across spatial grid for efficient batched inner product computation
\end{itemize}

\section*{Model Architecture for Plastic Deformation}

The Sequential DeepONet model used for the plastic deformation problem consists of a recurrent branch network that processes the time-dependent boundary displacement input, and a fully connected trunk network that encodes the spatial coordinates of the stress field. The detailed model configuration is given below.

\subsubsection*{Branch Network (Temporal Encoder)}
\begin{itemize}
    \item Type: GRU
    \item Input size: 1 (scalar displacement at each time step)
    \item Number of layers: 4
    \item Hidden units per layer: 256
    \item Dropout: 0.1
    \item Output size: 256 (final time-step hidden state)
    \item Normalization: Layer normalization applied to the final GRU output
\end{itemize}

\subsubsection*{Trunk Network (Spatial Encoder)}
\begin{itemize}
    \item Type: Fully Connected Network (FCN)
    \item Architecture: [2, 128, 128, 128, 256]
    \item Activation function: Tanh
    \item Dropout rate: 0.1 (applied after each hidden layer)
    \item Input: 2D spatial coordinates \( (x, y) \)
    \item Output: 256-dimensional latent vector per spatial location
\end{itemize}

\subsubsection*{Output Construction}
The final von Mises stress \( \hat{\sigma}(x, y) \) is predicted by computing the inner product between the latent temporal embedding from the branch network and the coordinate-dependent embedding from the trunk network:

\[
\hat{\sigma}(x_j, y_j) = \sum_{\ell=1}^{256} \phi_\ell(x_j, y_j) \cdot \psi_\ell(u(t)),
\]

where \( \psi \in \mathbb{R}^{256} \) is the GRU-encoded displacement history, and \( \phi(x_j, y_j) \in \mathbb{R}^{256} \) is the output of the trunk network at spatial location \( (x_j, y_j) \). A trainable bias is added to each output channel.

\subsubsection*{Model Summary}
\begin{itemize}
    \item Total number of trainable parameters: \textbf{1,450,113}
    \item Number of outputs: 1 (scalar von Mises stress per spatial node)
    \item Network supports efficient batch evaluation across time histories and spatial locations
\end{itemize}

\newpage

\section*{Model Architecture for Cosmic Radiation Dose}
The Sequential DeepONet model for the cosmic radiation dose prediction task is designed to learn the mapping from 7-day sequences of neutron monitor readings to the global spatial distribution of effective dose rates. The model processes multivariate temporal inputs via a recurrent branch network and spatial coordinates via a fully connected trunk network. Detailed architecture specifications are provided below.

\subsection*{Branch Network (Temporal Encoder)}
\begin{itemize}
    \item Type: GRU
    \item Input size: 12 (daily counts from 12 neutron monitor stations)
    \item Sequence length: 7 days
    \item Number of layers: 4
    \item Hidden units per layer: 256
    \item Dropout: 0.1
    \item Output size: 256 (final hidden state after the sequence)
    \item Normalization: Layer normalization applied to the final GRU output
\end{itemize}

\subsection*{Trunk Network (Spatial Encoder)}
\begin{itemize}
    \item Type: Fully Connected Network (FCN)
    \item Architecture: [2, 256, 256, 256, 256]
    \item Activation function: ReLU
    \item Dropout rate: 0.1 (applied after each hidden layer)
    \item Input: Spatial coordinates \( (x, y) \in \mathbb{R}^2 \)
    \item Output: 256-dimensional latent representation at each spatial location
\end{itemize}

\subsection*{Output Construction}
The model output, corresponding to the effective dose rate \( \hat{D}(x, y) \), is computed through an inner product between the latent representations from the branch and trunk networks:

\[
\hat{D}(x_j, y_j) = \sum_{\ell=1}^{256} \phi_\ell(x_j, y_j) \cdot \psi_\ell(N^{(7)}),
\]

where \( \psi \in \mathbb{R}^{256} \) encodes the neutron monitor sequence \( N^{(7)} \in \mathbb{R}^{7 \times 12} \), and \( \phi(x_j, y_j) \in \mathbb{R}^{256} \) encodes the spatial location. A learnable bias term is included in the final output.

\subsection*{Model Summary}
\begin{itemize}
    \item Total number of trainable parameters: \textbf{1,590,273}
    \item Number of outputs: 1 (scalar effective dose per spatial location)
    \item Designed for inference over 65,341 spatial grid points
\end{itemize}

\end{document}